\documentclass[letterpaper]{article} % DO NOT CHANGE THIS
\usepackage{aaai2026}  % DO NOT CHANGE THIS
\usepackage{times}  % DO NOT CHANGE THIS
\usepackage{helvet}  % DO NOT CHANGE THIS
\usepackage{courier}  % DO NOT CHANGE THIS
\usepackage[hyphens]{url}  % DO NOT CHANGE THIS
\usepackage{graphicx} % DO NOT CHANGE THIS
\usepackage{natbib}  % DO NOT CHANGE THIS AND DO NOT ADD ANY OPTIONS TO IT
\usepackage{caption} % DO NOT CHANGE THIS AND DO NOT ADD ANY OPTIONS TO IT
\usepackage{algorithm}
\usepackage{algorithmic}

\usepackage{newfloat}
\usepackage{listings}
\DeclareCaptionStyle{ruled}{labelfont=normalfont,labelsep=colon,strut=off} % DO NOT CHANGE THIS
\floatstyle{ruled}
\newfloat{listing}{tb}{lst}{}
\floatname{listing}{Listing}
\usepackage{amsmath,amssymb}
\usepackage{multirow}
\usepackage{booktabs}
\nocopyright %-- Your paper will not be published if you use this command
\title{RFKG-CoT: Relation-Driven Adaptive Hop-count Selection and Few-Shot Path Guidance for Knowledge-Aware QA}
\author{
    Chao Zhang,
    Minghan Li\thanks{Corresponding author},
    Tianrui Lv,
    Guodong Zhou
}
\affiliations{
    School of Computer Science and Technology, Soochow University, China\\

    czhang1@stu.suda.edu.cn,
    mhli@suda.edu.cn,
    trlvtrl@stu.suda.edu.cn,
    gdzhou@suda.edu.cn

}

\usepackage{bibentry}
\begin{document}

\maketitle

\begin{abstract}
Large language models (LLMs) often generate hallucinations in knowledge-intensive QA due to parametric knowledge limitations. While existing methods like KG-CoT improve reliability by integrating knowledge graph (KG) paths, they suffer from rigid hop-count selection (solely question-driven) and underutilization of reasoning paths (lack of guidance). To address this, we propose RFKG-CoT: First, it replaces the rigid hop-count selector with a relation-driven adaptive hop-count selector that dynamically adjusts reasoning steps by activating KG relations (e.g., 1-hop for direct ``brother" relations, 2-hop for indirect ``father-son" chains), formalized via a relation mask. Second, it introduces a few-shot in-context learning path guidance mechanism with CoT (think) that constructs examples in a ``question-paths-answer" format to enhance LLMs' ability to understand reasoning paths. Experiments on four KGQA benchmarks show RFKG-CoT improves accuracy by up to 14.7 pp (Llama2-7B on WebQSP) over KG-CoT. Ablations confirm the hop-count selector and the path prompt are complementary, jointly transforming KG evidence into more faithful answers.
\end{abstract}

\section{Introduction}

In recent years, large language models (LLMs) such as the GPT series \citep{achiam2023gpt} and Llama series have made breakthrough progress in natural language processing, excelling in text generation, question answering (QA), and logical reasoning \citep{wei2022chain}. However, in knowledge-intensive tasks, LLMs still face two core challenges: first, hallucination—due to limited reliable knowledge, LLMs sometimes struggle to generate credible answers based on black-box parametric knowledge \citep{ji2023survey}; second, factual staleness—static parameters make it hard to obtain real-time domain knowledge \citep{wang2023augmenting}. These issues severely restrict applications in high-accuracy scenarios like medical diagnosis and legal consulting.

To address the above problems, researchers have proposed knowledge enhancement methods, which provide factual support for LLMs through external knowledge sources (e.g., knowledge graphs, text corpora). Among them, knowledge graphs (KGs) have become an important knowledge enhancement tool due to their structured and interpretable characteristics. Their triple structure, composed of entities (nodes) and relations (edges), can clearly express semantic associations between entities \citep{bollacker2007freebase}. KG-based enhancement methods fall into two main categories: pre-training fusion, which embeds KG knowledge into LLM parameters (e.g., fusing KG information during pre-training \citep{zhao2024graph}), and reasoning-phase fusion, which dynamically introduces KG knowledge during inference (e.g., Retrieval-Augmented Generation (RAG) \citep{lewis2020retrieval}). As a technique to improve LLMs' reasoning ability, Chain-of-Thought (CoT) prompting significantly enhances model performance in logical reasoning tasks by guiding the model to generate intermediate reasoning steps and decomposing complex problems into simple subproblems \citep{wei2022chain}. However, traditional CoT relies solely on LLMs' parametric knowledge, still suffering from factual errors. To this end, KG-CoT has been proposed. It generates reasoning paths using structured KG knowledge and inputs these paths into LLMs as prompts to enable knowledge-aware reasoning \citep{zhao2024kg}.

\begin{figure}[t]
\includegraphics[width=\linewidth]{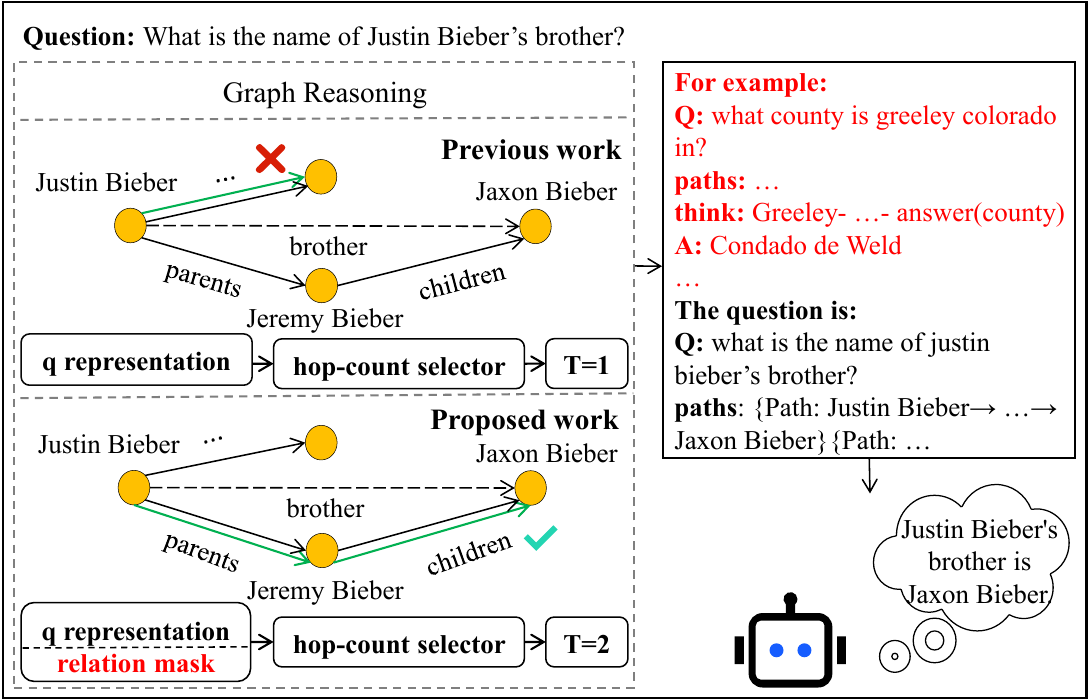}
\caption{Comparison of RFKG-CoT and KG-CoT, highlighting two key improvements: relation-driven hop-count selector (left red) and few-shot ICL path guidance (right red). The green arrows indicate the reasoning paths of different models.}
\label{fig: rfkg-cot}
\end{figure}

\citet{zhao2024kg} attempts to dynamically adjust reasoning steps using a graph reasoning model and hop-count selector. However, two critical limitations remain: (1) The hop-count selector underutilizes KG relational features, relying solely on question features to determine hops. For example, answering ``Who is Justin Bieber's brother?" requires 1 hop with a direct ``brother" relation but multiple hops with indirect ``father-son" chains (his father's other sons), yet the selector fails to adapt. (2) LLMs underutilize reasoning paths, as paths are input without guidance on interpretation, limiting the utilization of logical connections. 

To address these issues, this paper proposes Relation-Driven Adaptive Hop-count Selection and Few-Shot Path
Guidance for Knowledge-Aware QA (RFKG-CoT), an enhanced framework built on KG-CoT, as shown in Figure \ref{fig: rfkg-cot}. This paper makes two key contributions: First, we propose a relation-based dynamic reasoning step mechanism that enables reasoning steps to adaptively adjust path length based on entity relations in the knowledge base. Second, we introduce a few-shot in-context learning (ICL)  strategy \citep{dong2024survey} with CoT (think) to guide LLMs in better utilizing reasoning paths through examples, thereby enhancing knowledge-aware reasoning capabilities.

\section{Related Work}
The fusion of KGs and LLMs has emerged as a pivotal research direction in recent years, aiming to leverage KG-structured knowledge to compensate for LLMs' factual inaccuracies in knowledge-intensive tasks \citep{hu2023survey}. Based on the integration stage, such methods are categorized into two paradigms: pre-training fusion and reasoning-phase fusion. Pre-training fusion injects KG information during model pre-training to encode structured knowledge into LLM parameters. However, existing approaches (e.g., K-BERT \citep{liu2020k}, KG-BART \citep{liu2021kg}) suffer from two critical limitations: inapplicability to closed-source LLMs (e.g., GPT-4) due to parameter inaccessibility, and difficulty in updating dynamic KG knowledge (e.g., newly emerging entities) without costly re-pre-training. In contrast, reasoning-phase fusion dynamically introduces KG knowledge during LLM inference, offering superior flexibility and interpretability. Retrieval-Augmented Generation (RAG) \citep{lewis2020retrieval} is a representative method, which retrieves relevant triples from KGs, converts them into text, and feeds them to LLMs \citep{zhu-etal-2025-knowledge}. Recent advancements include integrating graph structure for precise subgraph retrieval \citep{mavromatis2024gnn} and iterative retrieval with LLM feedback \citep{yang2025beyond}.  

The challenge of dynamically determining the optimal reasoning depth in knowledge-intensive tasks represents a critical limitation in contemporary Chain-of-Thought (CoT) methodologies. Prior approaches largely fall into two categories: fixed-step reasoning and task-driven step selection. Traditional CoT frameworks \citep{wei2022chain} typically employ a predetermined number of reasoning steps, regardless of query complexity or underlying knowledge structure. This rigidity proves particularly suboptimal for KGs. KG-CoT \citep{zhao2024kg} introduced question-driven hop-count selection, dynamically setting hop counts based on question intent classification. However, its hop-count selection relies solely on the initial question embedding and lacks an explicit guidance mechanism for the reasoning path. This makes it difficult for LLMs to effectively utilize intermediate reasoning results, limiting their ability to reason with complex logic. We introduce a relation mask mechanism to fuse KG relational features with the question embedding to jointly generate the hop-count decisions.

Few-Shot Learning leverages models' pre-trained knowledge and strong contextual understanding, enabling rapid adaptation to new tasks (e.g., style-specific translation, text classification, code generation) with only a small number of examples. In KG-related tasks, few-shot is mainly used for relationship prediction and question answering. For instance, FewRel \cite{han2018fewrel} trains models to predict new relations through a small number of triple examples; KG-FewShotQA \cite{li2023few} constructs examples containing knowledge graph paths to guide LLMs to generate answers. However, such designs primarily focus on path-answer mapping and lack explicit components to guide how LLMs should interpret path elements or derive answers step-by-step from paths. To address this gap, we extend the path-based framework by fusing questions with reasoning paths to generate structured thinking chains, including a symbolic \textit{Think} template that maps path elements to step-by-step logic. This strengthens path-reasoning connections, enhancing LLMs' utilization of path semantics for answer derivation.

\section{Methodology}

\begin{figure*}[t]
\centering
\includegraphics[width=0.65\linewidth]{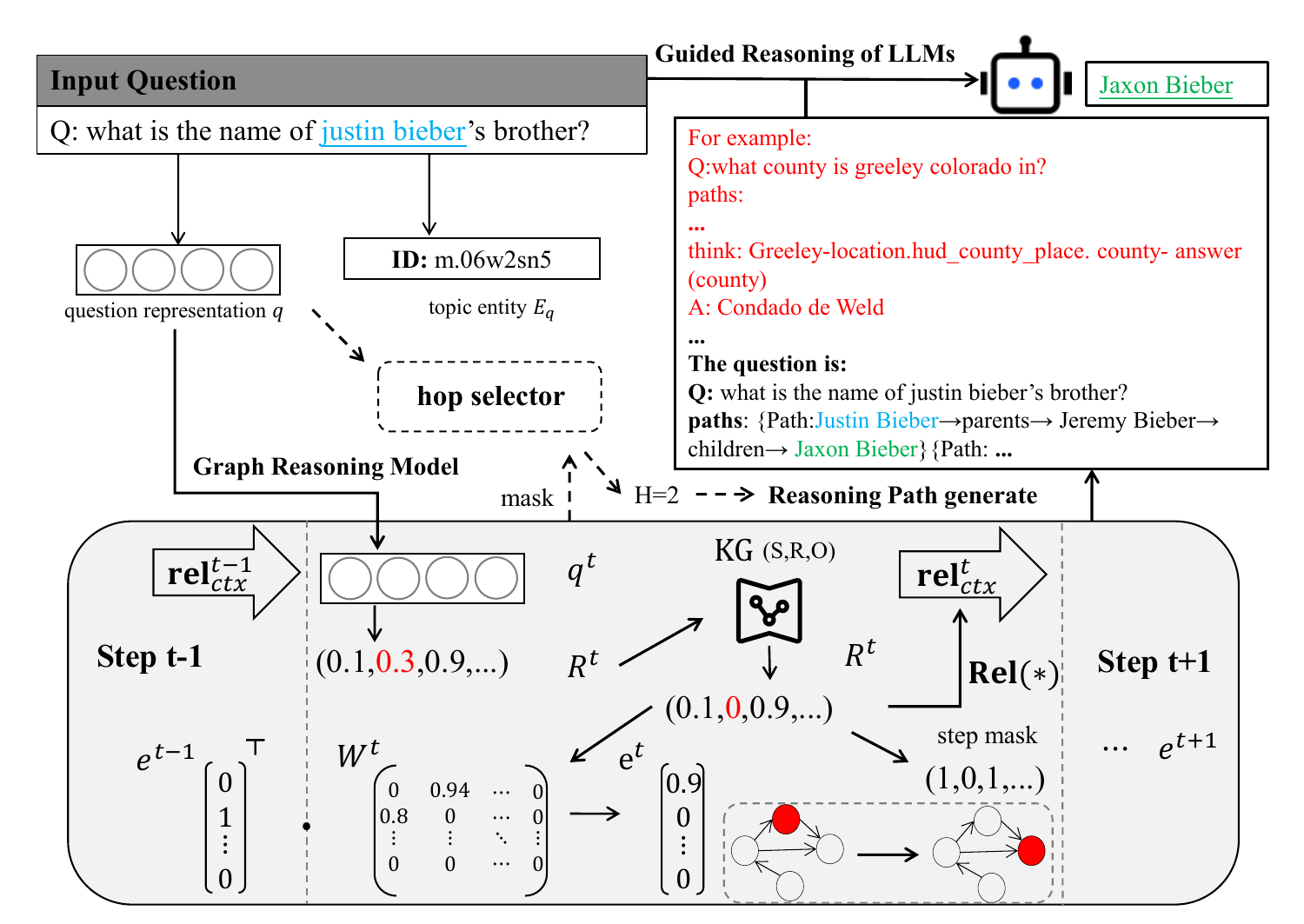}
\caption{Overview of RFKG-CoT. (1) A graph reasoning model performs reasoning over the knowledge graph, and a hop-count selector that incorporates both the question and knowledge graph relationships determines the required reasoning hops for answering the question. (2) Reasoning paths are generated via a dedicated path generation module. (3) The question is concatenated with generated paths, and selected few-shot examples are appended as prompts to input into the large language model (LLM) for final reasoning.}
\label{fig: overview}
\end{figure*}
RFKG-CoT first performs reasoning on the knowledge graph through a small-scale graph reasoning model, then generates reasoning paths. By collaborating with large language models, it combines the model's own knowledge and generated reasoning paths to derive the final result. Compared with the previous method \cite{zhao2024kg}, we have improved the graph reasoning model by optimizing the hop-count selection. 
We further provide few-shot exemplars that teach the LLM how to ground each path, ensuring it attends to the KG evidence.
The framework of RFKG-CoT is illustrated in Figure \ref{fig: overview}.

\subsection{Graph Reasoning Model}
We employ a graph reasoning model that decomposes problems into multiple steps via a hop-count selector, calculating relationship scores at each step to form executable multi-step reasoning paths on the knowledge graph.

\subsubsection{Initialization.}
We let $\mathcal{G}$ denote the knowledge graph, with $n$ representing the number of entities in the entity set and $m$ representing the number of relations in the relation set. The state of the topic entity in the problem is initialized using one-hot encoding as $\mathbf{e}^0 \in [0,1]^n$, where the position corresponding to the topic entity in the problem is set to 1 (and all others to 0). In addition, we initialize a triple matrix $\mathbf{M} \in [0,1]^{n \times n}$, where $M_{ij}=k$ indicates that the relation index between entity $i$ and entity $j$ is $k$ (e.g., $k$=3 maps to ``law.judge.cases" in WebQSP). Finally, we define the golden answer to the question as a one-hot vector 
$\mathbf{a}\in[0,1]^n$, that indicates which entity is the answer.

\subsubsection{Relation Score Calculation.}
We divide the graph reasoning process into $T$ steps (determined by the dataset). Starting from the topic entity, at step t ($t<T$), we calculate the scores for all relations in $\mathcal{G}$ based on the problem information focused on at the current stage, as $\mathbf{R}^t\in[0,1]^m$. The score of each relation $r^t_i\in \mathbf{R}^t$ represents its activation probability, and the model transmits entity scores through these activated relations. The calculation of $\mathbf{R}^t$ is as follows:
\begin{equation}
\mathbf{R}^t = Sigmoid(MLP_{KG}(\mathbf{q}^t)),
\end{equation}
where $MLP_{KG}$ converts the 768 dimensional question embedding $\mathbf{q}^t$ to an m dimensional vector of relation scores, and $\mathbf{q}^t$ denotes the query representation at step $t$. Since the model's focus on key parts of the question context may shift across different reasoning stages, at step t, we focus on a specific part of the question, and the calculation of the question representation $\mathbf{q}^t$ is as follows:
\begin{equation}
    \mathbf{q},(h_1, \cdots, h_{|q|}) =  Encoder(q),
\end{equation}
\begin{equation}
    \mathbf{Q}_t = f^t([\mathbf{q}, \mathbf{rel}_{ctx}]),
    \label{eq: relation_extraction}
\end{equation}
\begin{equation}
    \mathbf{b}^t = Softmax(\mathbf{Q}^t[h_1; \cdots ; h_{|q|}]),
\end{equation}
\begin{equation}
    \mathbf{q}^t = \sum_{i=1}^{|q|}b_i^t h_i,
\end{equation}
where $\mathbf{q}$ denotes the question embedding, and $(h_1, \cdots, h_{|q|})$ is the hidden state sequence of the question. The function $f^t$ is used to project the extracted question information and the relation information ($\mathbf{rel}_{ctx}$ initialized as a zero vector with the same shape as $\mathbf{q}^t$ at early steps) from the previous step onto the attention query $\mathbf{Q}$ at step $t$. Then, we calculate the attention weight for each word, and obtain the question representation at step t by performing a weighted sum of these weights. 

\subsubsection{Reasoning.}
We define a state transition matrix $\mathbf{W}^t\in [0,1]^{n \times n}$ based on relation scores $\mathbf{R}^t$:
\begin{equation}
    W^t_{ij} = 
    \begin{cases}
         R^t_k &\begin{aligned}[t] k=M_{ij},R^t_k \in \mathbf{R}^t,M_{ij}\in \mathbf{M}, \end{aligned}\\
        0      & Otherwise, 
    \end{cases}
\end{equation}
where $k$ is the relation index between entities $i$ and $j$ in the triple matrix $\mathbf{M}$, and $R^t_k$ is the score of relation $k$ at step $t$. The reasoning process propagates entity states via:
\begin{equation}
    e^t_j = \sum_{i=1}^{n}e_i^{t-1} \times W^t_{ij}.
\end{equation}
$\mathbf{e}^{t-1}$ denotes the entity state vector at step $t-1$. This operation diffuses entity probabilities along 1-hop neighbors weighted by relation scores.

\begin{algorithm}[tb]
\caption{Relation Activation Mask Generation}
\label{alg: relactive}
\textbf{Input}: Triple matrix $\mathbf{M}$, relation score $\mathbf{R}^t$.\\
\textbf{Initialize}: $\mathbf{mask} \in \{0,1\}^m\gets 0$.\\
\textbf{Output}: Relation activation mask $\mathbf{mask}$.  \\
\begin{algorithmic}[1] %[1] enables line numbers
\FOR{$t = 1, \cdots, T$}
\STATE Initialize step-wise mask $\mathbf{mask}_{i} \in \{ 0,1 \}^m \gets \mathbf{0}$.
\STATE Get triple indices $\mathbf{I}_t = \{ j \mid obj\_p[j] > 0 \}$.
\STATE Extract relation index  $\mathbf{I}_r = \{ M_{i,j} \mid j \in \mathbf{I}_t, i \in sub\_p[j] \}$ (relations from subject entities $i$ to target entities $j$).
\STATE Activate step-wise mask $\mathbf{mask}_i[k] = 1$ for all $k \in \mathbf{I}_r$.
\STATE Filtered relation scores $\mathbf{R}^t \gets \mathbf{R}^t \odot \mathbf{mask}_i$.
\STATE Update global mask $\mathbf{mask} = \mathbf{mask} \lor (\mathbf{R}^t > 10^{-6})$.
\ENDFOR
\STATE \textbf{return} $\mathbf{mask}$
\end{algorithmic}
\end{algorithm}

\subsubsection{Adapted Hop-Count Selection.}
Unlike the static hop-count selector proposed in \citet{zhao2024kg} that determines hops solely based on question features, we propose a relation-driven mechanism that adapts reasoning steps to the KG structure. Specifically, we introduce a relation activation mask $\mathbf{mask} \in \{ 0,1 \}^m$ (initialized to all zeros) to track which relations are activated during reasoning, as detailed in Algorithm~\ref{alg: relactive}:

In each of the $T$ reasoning steps, we first identify activated relations via non-zero target entity probabilities (stored in $\mathbf{obj\_p}$, where $obj\_p[i]$ denotes the probability of the target entity in the $i$-th triple, computed as the product of subject probability $\mathbf{sub\_p}$ and relation probability $\mathbf{rel\_p}$ for the triple), update a step-wise mask to filter the current relation scores $\mathbf{R}^t$, and accumulate valid activations into the  $\mathbf{mask}$ (retaining relations with scores exceeding $10^{-6}$). The filtered relation scores are then used to extract relation information (via $Rel(*)$) for the next step of question encoding (see Eq. \eqref{eq: relation_extraction}), enabling the model to incorporate KG structure into reasoning adaptively. 
\begin{equation}
    \mathbf{rel}_{ctx} = Rel(\mathbf{R}^t).
\end{equation}

After $T$ steps, the $\mathbf{mask}$ captures all relations that contributed to the reasoning process. We combine this mask with the original question $\mathbf{q}$ to determine the optimal number of hops $H$:
\begin{equation}
    \mathbf{c} = Softmax(MLP_{T}([\mathbf{q},\mathbf{mask}])),
\end{equation}
where $MLP_{T}$ converts the 768+m dimensional concatenated vector of $\mathbf{q}$ and $\mathbf{mask}$ to $T$ dimensional vector,
\begin{equation}
    H = \arg\max_{t \in \{1, 2, \cdots, T\}} c_t,
\end{equation}

and subsequently compute the final entity score $\overline{e}$:
\begin{equation}
\overline{\mathbf{e}}=\sum_{t=1}^{T}c_t \mathbf{e}^t.
\end{equation}

\subsubsection{Training.}
We optimize the graph reasoning model by computing the $\mathbf{L}2$ distance between $\overline{\mathbf{e}}$ and $\mathbf{a}$:
\begin{equation}
    \mathcal{L}=\lVert\mathbf{\overline{e}}-\mathbf{a}\rVert^2.
\end{equation}

\begin{algorithm}[tb]
\caption{Generate reasoning paths}
\label{alg: generate paths}
\textbf{Input}: Input question q, retrieved knowledge subgraph $\mathcal{G}$.\\
\textbf{Initialize}: Entity score $\mathbf{e}^0 \gets$ extract topic entity entity $\mathbf{E}_q$, triplet matrix $\mathbf{M} \gets$ extract triples from $\mathcal{G}$.\\
\textbf{Output}: Reasoning paths \\
\begin{algorithmic}[1]
\FOR{$t = 1,...T$}
\STATE Compute the question representation $\mathbf{q}^t$ using (2)-(5).
\STATE Compute the relation score $\mathbf{R}^t$ using (1).
\STATE Update relation activation mask $\mathbf{mask}$ using algorithm ~\ref{alg: relactive}.
\STATE compute transition matrix $\mathbf{W}^t$ using (6).
\STATE Entity score $\mathbf{e}^t \gets$ reasoning using (7).
\STATE Compute the relation information $\mathbf{rel}_{ctx}$ using (8).
\ENDFOR
\STATE Compute final scores $\overline{\mathbf{e}}$ using (9), (11) and select top-$K$ entities $\mathbf{E}^K$.
\STATE Initialize $L_m \gets$ topic entity $E^0$.
\FOR{$t = 1,...H$}
\STATE Extract t-hop paths $\mathbf{P}^t$ using $\mathbf{W}^t$ and paths in $L_m$.
\STATE Update intermediate path list $L_m$ with $\mathbf{P}^t_{qj}$.
\IF{Object entity $\mathbf{E}_j \in \mathbf{E}^K$}
\STATE Update reasoning path list $L_c$ with $\mathbf{P}^t_{qj}$.
\ENDIF
\ENDFOR
\STATE Select $N$ paths for each top-$K$ entity $\mathbf{E}^K$ from $L_c$.
\STATE \textbf{return} list of reasoning paths
\end{algorithmic}
\end{algorithm}

\subsection{Generation of Reasoning Paths}
In the path generation phase, two lists $L_c$ and $L_m$ are maintained to store candidate reasoning paths and intermediate reasoning paths, respectively. Starting from the topic entity, we identify the predicted entities for each sample via the hop-count selector, select the top-$K$ answer entities, and then generate reasoning paths from the topic entity to these top-$K$ predicted entities using the transition matrix $W^1, W^2, \cdots, W^T$, as shown in Algorithm~\ref{alg: generate paths}. $\mathbf{P}^t_{qj}$ denotes the path starting from the topic entity to the target entity. Each answer entity may correspond to multiple candidate paths in $L_c$. We calculate the average of the relationship scores in each path and take it as the final path score.

\subsection{Guided Reasoning of LLMs}
For the top-$K$ candidate entities with high confidence scores, we extract the path with the highest path score for each entity. Consequently, we generate $K$ reasoning paths per question via the reasoning path generation method.

We use `$\to$' to connect entities and relations, and a 1-hop reasoning path is represented as:

\begin{equation}
 \mathbf{E}_i \to \mathbf{Rel}_{ij} \to \mathbf{E_j}.
\end{equation}

\begin{figure}[h]
\centering
\includegraphics[width=0.98\columnwidth]{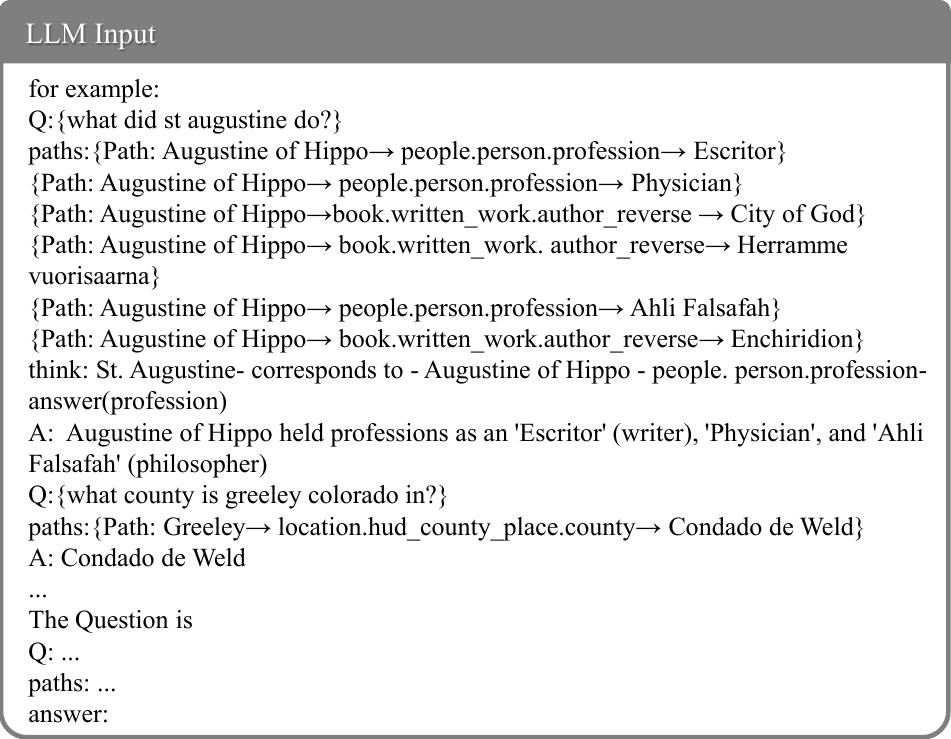}
\caption{Few-shot prompt template used in the LLM stage.}
\label{fig: prompt}
\end{figure}

We serialize the $K$ reasoning paths and concatenate them with the question context to form the final input sequence. To enhance LLMs' ability to utilize reasoning paths, we introduce a manually curated few-shot prompt containing three fixed examples. Each example follows a structured ``Question-Paths-Answer" format with four essential components: the original natural language query, serialized KG paths in the form of $\mathbf{E}_i \to \mathbf{Rel}_{ij} \to \mathbf{E_j}$, a symbolic reasoning template (Think) mapping path elements to answer constraints, and explicit answer format constraints. The prompt template is shown in Figure ~\ref{fig: prompt}.
As shown, we prepend few-shot ICL exemplars with CoT (think). They calibrate the model’s attention to the critical fields of each path (entity, relation, direction), thereby reducing free-form hallucination. The later ablation study confirms that removing the exemplars drops accuracy.

\section{Experiments}
\subsection{Datasets}
We evaluate RFKG-CoT on four widely used KGQA benchmarks, covering single-hop, multi-hop, and open-domain scenarios to test its knowledge-aware reasoning capabilities comprehensively:
\begin{itemize}
\item \textbf{WebQSP} \cite{yih2016value}: Multi-hop QA benchmark (1-2 hops) based on Freebase with compositional questions.
\item \textbf{CompWebQ} \cite{talmor2018web}: Extension of WebQSP with increased complexity and constraints. 
\item \textbf{SimpleQuestions} \cite{petrochuk2018simplequestions}: Single-hop QA requiring direct KG lookups.
\item \textbf{WebQuestions} \cite{berant2013semantic}: Open-domain QA from real user queries.
\end{itemize}
Following the method of \citet{zhao2024kg}, we construct knowledge subgraphs via neighborhood retrieval:
\begin{itemize}
    \item \textbf{SimpleQuestions}: 1-hop neighborhood of topic entities.
    \item \textbf{CompWebQ}: We adopt the method of \citet{shi2021transfernet} to retrieve subgraphs for each question.
    \item \textbf{WebQSP \& WebQuestions}: Bidirectional 2-hop expansion with reverse relations added to enhance connectivity.
\end{itemize}
The detailed information of the datasets is shown in Table~\ref{dataset}.

\begin{table}[h]
\centering
\begin{tabular}{lrrr}
\toprule
\textbf{Dataset} & \textbf{Total} & \textbf{Train} & \textbf{Test} \\
\midrule
WebQSP & 4637 & 2998 & 1639 \\
CompWebQ & 31154 & 27623 & 3531 \\
SimpleQuestions & 97597 & 75910 & 21687 \\
WebQuestions & 5810 & 3778 & 2032 \\
\bottomrule
\end{tabular}
\caption{Dataset statistics.}
\label{dataset}
\end{table}

\subsection{Baselines}
\subsubsection{Prompting Baselines.}
We compare with:
\begin{itemize}
    \item \textbf{IO} \cite{patel2023bidirectional}: Direct input-output prompting.
    \item \textbf{CoT} \cite{wei2022chain}: Chain-of-thought prompting.
    \item \textbf{SC} \cite{wang2022self}: Self-consistency with multiple reasoning paths.
\end{itemize}

\subsubsection{Retrieval-Augmented Baselines.}
We selected retrieval-augmented methods for each benchmark, including GNN-RAG \cite{mavromatis2024gnn} that leverages graph networks for context-aware retrieval, DiFaR \cite{baek2023direct} with a dynamic factual alignment mechanism, CBR \cite{das2021case} utilizing analogical reasoning over knowledge snippets, and FiE \cite{kedia2022fie} that integrates knowledge through encoder modifications.

\subsubsection{Knowledge Base Question Answering Baselines.}
We compare with representative traditional knowledge base question answering models (non-LLM-based) including UniKGQA \cite{jiang2023unikgqa} and RNG \cite{ye2022rng}.

\subsubsection{LLM+KG Baseline.}
We take KG-CoT \citep{zhao2024kg} as the key LLM+KG baseline, as it integrates knowledge graphs with chain-of-thought reasoning. Our RFKG-CoT builds upon KG-CoT with two critical improvements: During graph reasoning, we introduce relation activation information to optimize hop-count selection, enabling the model to flexibly handle semantically equivalent information by prioritizing relevant relations. We incorporate few-shot examples during LLM reasoning to enhance the model’s ability to utilize extracted reasoning paths.

\subsection{Implementation Setting}
We incorporated reversed relations into the relation graph. For experimental settings, the number of reasoning steps T was set to 2 for WebQSP and WebQuestions, while for SimpleQuestions and CompWebQ, it was set to 1 and 3, respectively. In all experiments, we used the pre-trained `bert-base-uncased' model \citep{devlin2019bert} as the question encoder and fine-tuned its parameters.  We trained the graph reasoning model using the RAdam optimizer for 60 epochs with a learning rate of 1e-3 (for graph layers) and 1e-5 (for BERT parameters). For large language models, we conducted evaluations by calling ChatGPT and GPT-4 via the OpenAI API, selecting ``gpt-3.5-turbo" and ``gpt-4" as the backbones of the large language models and using the default settings of the OpenAI API. Specifically, for each question, we generated one reasoning path for each of the top-10 candidate answer entities, and used these paths together with the question as input to enable the large language models to generate answers for evaluation directly.

\subsection{Main Results}

\begin{table*}[htbp]
\small
\centering
\begin{tabular}{lccccc}
\toprule
\multirow{2}{*}{\textbf{Model}} &
\multirow{2}{*}{\textbf{AccessKB}} &
\multicolumn{2}{c}{\textbf{Multi-hop QA}} &
\textbf{Single-hop QA} & 
\textbf{Open-domain QA} \\ 
\cmidrule{3-6}
&&   \textbf{WebQSP} & \textbf{CompWebQ} & \textbf{SimpleQuestions} & \textbf{WebQuestions} \\
\midrule
ChatGPT+IO \citep{patel2023bidirectional} & $\times$ & 63.3 & 37.6 & 20.0 & 48.7 \\
ChatGPT+CoT \citep{wei2022chain} & $\times$ & 62.2 & 38.8 & 20.3 & 48.5 \\
ChatGPT+SC \citep{wang2022self} & $\times$ & 61.1 & 45.4 & 18.9 & 50.3 \\
\midrule
Previous RA  & $\checkmark$ & 90.7$^{\alpha}$ & \textbf{70.4}$^{\beta}$ & 85.8$^{\gamma}$ & 56.3$^{\delta}$ \\
Previous KBQA  & $\checkmark$ & 76.6$^{\varepsilon}$ & 52.2$^{\varepsilon}$ & 71.1$^{\eta}$ & -- \\
ChatGPT+KG-CoT \citep{zhao2024kg} & $\checkmark$ & 82.1 & 51.6 & 77.8 & 66.5 \\
GPT-4+KG-CoT \citep{zhao2024kg} & $\checkmark$ & 84.9 & 62.3 & 86.1 & 68.0 \\
\midrule
\textbf{ChatGPT+RFKG-CoT (ours)} & $\checkmark$ & 89.9 & 61.4 & 79.2 & \textbf{77.0} \\
\textbf{GPT-4+RFKG-CoT (ours)} & $\checkmark$ & \textbf{91.5} & 65.1 & \textbf{87.0} & \textbf{78.2} \\
\bottomrule
\end{tabular}

\caption{
Accuracy Comparison between RFKG-CoT and Other Baseline Methods(e.g., $\alpha$: GNN-RAG \cite{mavromatis2024gnn}, $\beta$: CBR \cite{das2021case}, $\gamma$: DiFaR \cite{baek2023direct}, $\delta$: FiE \cite{kedia2022fie}, $\varepsilon$: UniKGQA \cite{jiang2023unikgqa}, and $\eta$: RNG \cite{ye2022rng} ) on Multi-hop QA, Single-hop QA, Open-domain QA.
}
\label{tab: main}
\end{table*}

\begin{table}[ht]
\centering
\begin{tabular}{lcc}
\toprule
\textbf{Method} & \textbf{WebQSP} & \textbf{CompWebQ} \\
\midrule
%  Llama2-7B
\multicolumn{3}{c}{Llama2-7B} \\
\midrule
KG-CoT   & 72.4  & 46.7  \\
RFKG-CoT & \textbf{87.1}    & \textbf{51.9} \\
$\Delta$ & (+14.7) & (+5.2) \\
\midrule
% Llama2-13B
\multicolumn{3}{c}{Llama2-13B} \\
\midrule
KG-CoT   & 74.6  & 50.0  \\
RFKG-CoT & \textbf{88.7}    & \textbf{55.5} \\
$\Delta$ & (+14.1) & (+5.5) \\
\midrule
% ChatGPT
\multicolumn{3}{c}{ChatGPT} \\
\midrule
KG-CoT   & 82.1  & 51.6  \\
RFKG-CoT & \textbf{89.9}   & \textbf{61.4} \\
$\Delta$ & (+7.8) & (+9.8) \\
\midrule
% GPT-4
\multicolumn{3}{c}{GPT-4} \\
\midrule
KG-CoT   & 84.9  & 62.3  \\
RFKG-CoT & \textbf{91.5}  & \textbf{65.1} \\
$\Delta$ & (+6.6) & (+2.8) \\
\bottomrule
\end{tabular}
\caption{Performance comparison (\%) on WebQSP and CompWebQ across LLM architectures.}
\label{tab: llms}
\end{table}

As shown in Table \ref{tab: main}, RFKG-CoT achieves strong performance across knowledge-intensive QA benchmarks. On multi-hop QA tasks, it performs well on WebQSP (91.5\%) and WebQuestions (78.2\%), surpassing GPT4+KG-CoT by 6.6 and 10.2 points, respectively. For CompWebQ, the GPT-4 variant reaches 65.1\%, significantly outperforming KG-CoT (62.3\%) despite subgraph coverage limitations. In single-hop scenarios, it performs strongly on SimpleQuestions (87\%), with the modest gain partly attributed to two factors: the hop-count selector (designed for multi-hop tasks) has limited utility here, and entity IDs in generated paths (lacking semantic alignment) may have constrained further improvements. Notably, when evaluating only answerable CompWebQ queries (2,611 samples), RFKG-CoT reaches 88.1\% accuracy, which is a 3.8 pp improvement over KG-CoT (84.3\%), demonstrating its robustness to subgraph incompleteness. This confirms that performance gaps primarily stem from knowledge coverage constraints rather than limitations in reasoning capability.

\begin{table}[ht]
\small
\centering
\begin{tabular}{lcc}
\toprule
    Method & WebQSP & CompWebQ  \\
\midrule
    KG-CoT (Baseline) & 82.1 & 51.6 \\
    + Relation Activation Mask Only & 85.5 & 59.8 \\
    + Few-Shot Guidance Only & 87.7 & 57.8 \\
\midrule
    \textbf{RFKG-CoT (Full)} & 89.9 & 61.4 \\
\bottomrule
\end{tabular}
\caption{Ablation study on WebQSP and CompWebQ with ChatGPT for verifying improved components (Accuracy \%).}
\label{tab: ablation}
\end{table}

\subsection{Robustness across LLM Architectures}
To assess the generalizability of RFKG-CoT, we evaluate its performance with diverse LLMs spanning open-source (Llama2-7B, Llama2-13B) and commercial APIs (ChatGPT, GPT-4). As shown in Table \ref{tab: llms}, RFKG-CoT consistently enhances KG-CoT across all configurations with two key findings: First, we observe an inverse scaling relationship where smaller LLMs exhibit substantially larger relative gains (e.g., Llama2-13B achieves +14.1 pp on WebQSP and +5.5 pp on CompWebQ) compared to advanced models like GPT-4 (+6.6 pp/+2.8 pp). This pattern emerges because RFKG-CoT's path guidance compensates more effectively for parametric knowledge gaps in capacity-constrained models. Second, performance plateaus approach the intrinsic limits of our graph reasoning module: analysis shows the correct answer appears in generated reasoning paths for 92.3\% of WebQSP. Similar constraints apply to CompWebQ. Thus, while LLM scaling improves absolute performance, the remaining gap is primarily constrained by the knowledge grounding component rather than LLM capabilities.

\subsection{Ablation study}
To verify the effectiveness of the improved components, we conducted ablation studies on the WebQSP and CompWebQ datasets, using ChatGPT for evaluation. Table \ref{tab: ablation} compares the full model against KG-CoT and single-component variants. The ``Relation Activation Mask Only'' variant outperforms KG-CoT by 3.4 pp on WebQSP but 8.2 pp on CompWebQ. This larger gain on CompWebQ indicates the mask’s strength in improving path quality by prioritizing relevant relations, particularly valuable for more complex reasoning scenarios. In contrast, ``Few-Shot Guidance Only" yields balanced gains (+5.6 pp on WebQSP, +6.2 pp on CompWebQ), enhancing path utilization by teaching LLMs to interpret structured paths. Unlike the mask (optimizing the selection of reasoning hops), few-shot guidance optimizes how paths are used. The full RFKG-CoT achieves the highest accuracy (89.9\%/61.4\%). This synergy arises because high-quality paths (from the mask) are better utilized via few-shot guidance, forming a closed loop. Both components are critical for robust reasoning.

\subsection{Hyperparameter Sensitivity Analysis}
To investigate how key input parameters affect performance, we analyze the sensitivity of RFKG-CoT to two critical hyperparameters consistent with the settings in \citet{zhao2024kg} for direct comparability: $K$(number of candidate answer entities, tested values: 5, 10, 15) and $N$(reasoning paths per entity, tested values: 1, 5, 10).

\begin{figure}[ht]
\centering
\includegraphics[width=1.0\columnwidth]{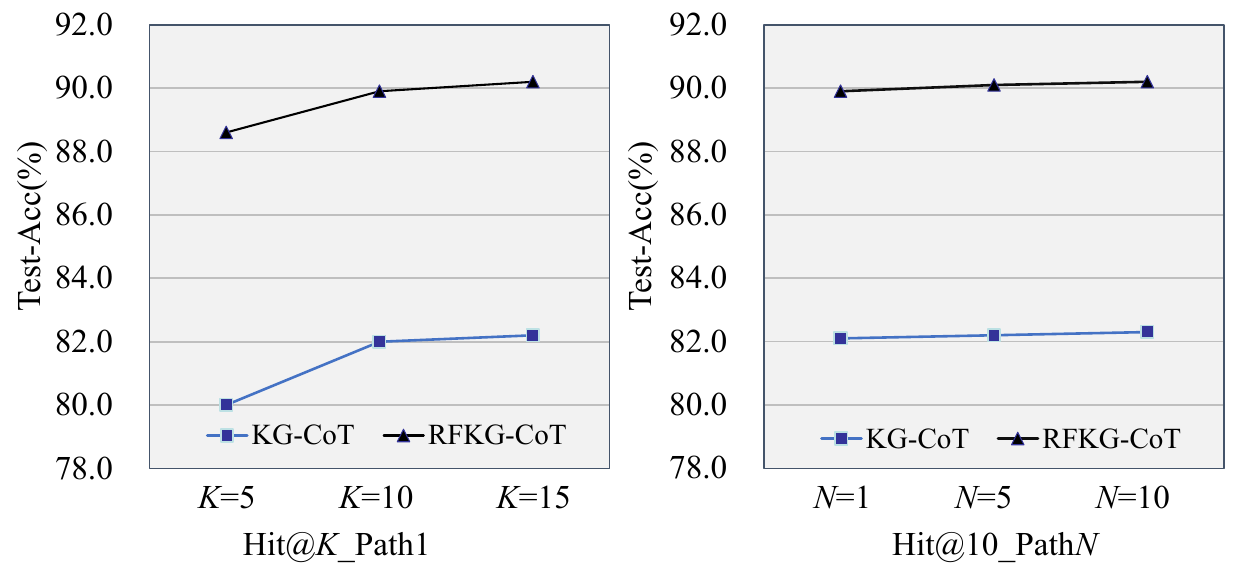} 
\caption{Effect of the number of candidate answer entities (left) and paths per entity (right) on performance in WebQSP.}
\label{fig: hyperparameter}
\end{figure}

Experiments are conducted on WebQSP using ChatGPT.
As shown in the left part of Figure \ref{fig: hyperparameter}, both models exhibit improved accuracy with increasing $K$, following a trend of ``early rapid gain, late saturation", which is consistent with KG-CoT’s observations. Expanding $K$ from 5 to 10 boosts performance for both models, driven by increased coverage of valid answer entities in reasoning paths. However, further increasing $K$ to 15 yields marginal gains, as lower-ranked candidates contribute minimally to reasoning due to reduced confidence and semantic relevance.  

The right part of Figure \ref{fig: hyperparameter} characterizes the effect of the number of reasoning paths per candidate answer entity. Consistent with KG-CoT’s findings, both models exhibit minimal performance gains as $N$ increases. This trend stems from a shared mechanism: paths are selected by confidence scores, where the top-ranked paths already capture the most critical reasoning signals. As $N$ increases, incremental paths inherently have lower confidence, contributing marginally to LLM inference.

\subsection{Impact of Few-Shot Example Quantity}
To investigate how the number of few-shot examples affects reasoning performance, we conducted a controlled experiment on WebQSP using ChatGPT+RFKG-CoT, with the baseline configuration $E=0$. We then systematically increased the number of examples from $E=1$ to $E=5$ to evaluate performance sensitivity to guidance quantity.

\begin{figure}[htbp]
\centering
\includegraphics[width=0.65\columnwidth]{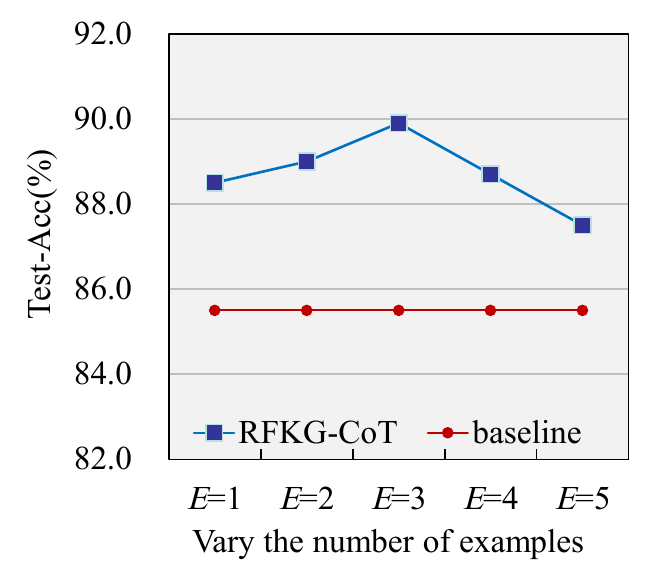} 
\caption{Accuracy sensitivity to few-shot example quantity on WebQSP (ChatGPT). The dashed line indicates the baseline (RFKG-CoT without few-shot examples).}
\label{fig: fewshot_quantity}
\end{figure}

As shown in Figure~\ref{fig: fewshot_quantity}, RFKG-CoT's accuracy exhibits a distinct non-monotonic relationship with $E$. The baseline $E=0$ demonstrates competent performance using only relation-driven path selection. Adding minimal guidance ($E=1$) yields substantial gains, with progressive improvement up to the $E=3$ optimum. Beyond this optimum, additional examples progressively degrade results, revealing a critical trade-off between guidance completeness and cognitive load. This pattern aligns with transformer architecture constraints, validating our default three-example design and confirming that calibrated few-shot guidance complements relation-driven path selection.

\section{Conclusion}

We presented RFKG-CoT, a knowledge-aware QA framework that fuses symbolic graph reasoning with LLMs. It improves from previous work like KG-CoT via
(1) a relation-driven hop selector that records activated relations with a mask and dynamically chooses the right reasoning depth, and
(2) a few-shot path prompt that guides the LLM to ground each retrieved path.

On WebQSP, CompWebQ, SimpleQuestions and WebQuestions, RFKG-CoT consistently outperforms prior work, boosting accuracy by up to 14.7 pp. Gains are largest on smaller open-source LLMs, underlining the value of explicit external knowledge. Ablations show the hop mask and path prompt each help independently and, when combined, reinforce one another by selecting better paths and exploiting them more effectively.

Beyond performance, RFKG-CoT produces interpretable reasoning traces and naturally supports incremental KG updates without retraining the LLM, making it attractive for real-world, rapidly evolving domains such as legal or biomedical search.

\section{Acknowledgments}
This work was supported by the National Natural Science Foundation of China (No. 62376178), and the Priority Academic Program Development of Jiangsu Higher Education Institutions.

\bibliography{main}

\end{document}